\documentclass{article} 
\usepackage{iclr2027_conference,times}
\usepackage{booktabs}
\usepackage{amsfonts}
\usepackage{amsmath}
\usepackage{amssymb}
\usepackage{nicefrac}
\usepackage{microtype}
\usepackage{xcolor}
\usepackage{colortbl}
\definecolor{pdmscol}{gray}{0.92}
\definecolor{defrow}{gray}{0.85}
\newcolumntype{P}{c}
\newcolumntype{Q}{c}
\usepackage{graphicx}
\usepackage{multirow}
\usepackage{wrapfig}
\usepackage{algorithm}
\usepackage{algorithmic}
\usepackage{subcaption}
\usepackage{pifont}
\usepackage{makecell}

\usepackage{amsmath,amsfonts,bm}

\def\eqref#1{equation~\ref{#1}}

\def\1{\bm{1}}

\DeclareMathAlphabet{\mathsfit}{\encodingdefault}{\sfdefault}{m}{sl}
\SetMathAlphabet{\mathsfit}{bold}{\encodingdefault}{\sfdefault}{bx}{n}

\usepackage{hyperref}
\usepackage{url}

\title{SimWAM: A Simple World Action Model for End-to-End Autonomous Driving}

\author{%
\textbf{Zongchuang Zhao$^{1}$,
Xin Zhou$^{1}$,
Tianyang Xu$^{1}$,
Zhengyang Sun$^{1}$}
\\
\textbf{
Kaixuan Zhou$^{2}$,
Yu Wu$^{2}$,
Honglin Li$^{2}$,
Dingkang Liang$^{1\dagger}$,
Xiang Bai$^{1}$}
\\ \\
$^1$Huazhong University of Science \& Technology, $^2$ Dongfeng Research \& Development Institute.
\\
\texttt{\{zcuangzhao, xzhou03, dkliang, xbai\}@hust.edu.cn}
}

\iclrfinalcopy 

\begin{document}

\maketitle
\lhead{Preprint}

\begingroup
\makeatletter
\let\thefootnote\relax\footnotetext{$\dagger$ Project leader. Work done during Zongchuang Zhao's internship at Dongfeng R\&D Institute.}
\endgroup

\begin{abstract}
In autonomous driving, World-Action Models (WAMs) have improved end-to-end planning by transferring video dynamics priors to action prediction, but many still couple planning with future-video generation at inference, incurring substantial computational overhead. We present \textbf{SimWAM}, a simple yet effective WAM that leverages future-video prediction solely as a training-time supervision signal. It co-trains a pretrained video expert and a lightweight action expert with joint flow matching. An isolated attention mask keeps action prediction independent of future frames, allowing trajectory prediction without future-frame generation at inference. This design supports multiple pretrained video backbones and independent action-expert scaling within a shared attention interface, while preserving the joint learning objective. Moreover, we apply reinforcement learning to optimize a compositional driving reward beyond trajectory imitation. Experiments show that SimWAM achieves $91.9$ PDMS on NAVSIM with a favorable trade-off between accuracy and latency among world-model-based planners, while transferring zero-shot to nuScenes. It also achieves competitive planning accuracy on WOD-E2E and PhysicalAI-Autonomous-Vehicles. These results position SimWAM as a plain yet solid baseline for efficient autonomous driving. The code and model weights are available at \url{https://github.com/H-EmbodVis/SimWAM/}.
\end{abstract}

\section{Introduction}

End-to-end autonomous driving (AD)~\citep{codevilla2018end} maps raw sensor observations directly to a planned trajectory with a unified network. Joint optimization removes hand-crafted interfaces and reduces error propagation in the classical perception, prediction, and planning pipeline~\citep{pomerleau1988alvinn,bojarski2016end}. Although recent end-to-end planners~\citep{hu2023planning,jiang2023vad,liao2025diffusiondrive} have steadily improved planning accuracy, they remain largely black-box mappings from observations to trajectories. In such models, traffic semantics, user intent, and scene dynamics remain only implicitly encoded within the learned mapping.

Vision-Language-Action (VLA) models~\citep{kim2024openvla,black2024pi0,chen2026recogdrive,li2026sgdrive,fu2025orion} address the semantic limitation by adapting pretrained vision-language models to driving. Their world knowledge and reasoning improve scene understanding and connect trajectory generation with user intent. Many driving VLAs~\citep{zhao2025cot,zhou2026autovla,wang2025alpamayo} further produce an explicit rationale before trajectory prediction, improving interpretability in complex and instruction-conditioned scenarios. Recent methods~\citep{zeng2026futuresightdrive,tan2026latent,peng2026colavla} introduce future-scene generation or latent reasoning to strengthen spatiotemporal understanding. However, these components remain loosely coupled with action prediction and often require additional training stages or sequential inference. Motion and temporal evolution are still modeled only indirectly, calling for an explicit prior over world dynamics.

World models meet this demand by furnishing an explicit prior over how the environment evolves under motion. Building on this principle, recent World-Action Models (WAMs) in embodied \begin{wrapfigure}{r}{0.45\textwidth}
  \centering
  \includegraphics[width=0.44\textwidth]{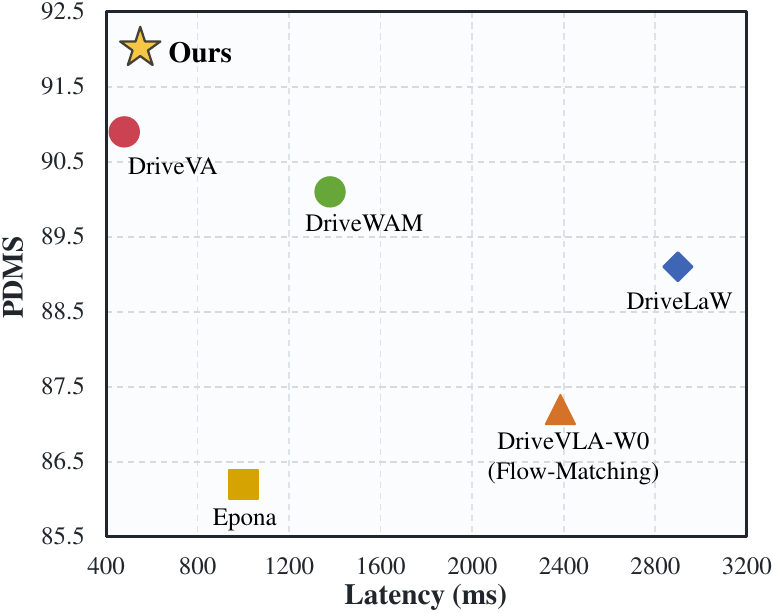}
  \caption{SimWAM achieves the best PDMS among world-model-based planners, with a favorable trade-off between accuracy and latency on one A100 GPU.}
  \label{fig:intro}
  \vspace{-10pt}
\end{wrapfigure}intelligence, such as DreamZero~\citep{ye2026world} and LingBot-VA~\citep{li2026causal}, jointly predict future observations and actions through pretrained video-generation backbones. This world-action paradigm has recently been adopted in autonomous driving. DriveLaW~\citep{xia2026drivelaw} and DriveWAM~\citep{shi2026drivewam} co-train a video predictor and a planner, allowing anticipated scene dynamics to inform trajectory generation. Nevertheless, existing driving WAMs commonly follow an \emph{imagine-then-act} pipeline in which the planner conditions its output on explicit future visual prediction. This reliance on future synthesis incurs substantial inference overhead, as reflected in Fig.~\ref{fig:intro}.

Indeed, Fast-WAM~\citep{yuan2026fast}, a world-action model from embodied intelligence, shows that explicit future synthesis is unnecessary for effective world-action learning, as video co-training benefits action prediction primarily through \emph{training-time} representation learning rather than \emph{test-time} imagination. Building on this insight, we introduce \textbf{SimWAM}, a plain yet effective WAM that uses video generation as a training signal while retaining direct trajectory prediction at inference. SimWAM jointly trains a pretrained video expert and a lightweight action expert with flow matching. A simple isolated attention mask prevents the action expert from accessing future frames, so inference bypasses explicit future-frame prediction while the action expert still leverages the learned traffic-dynamics prior. Building on this mask, more advanced video generators could further improve the learned prior without touching the action expert or the inference pipeline. Furthermore, we reformulate the deterministic flow ODE as a stochastic SDE and reinforce the action expert with GRPO~\citep{guo2025deepseek,liu2026flow}, enabling diverse maneuver exploration and direct optimization of a compositional driving reward. Rather than claiming algorithmic superiority, this work establishes a simple and solid WAM baseline for exploring the potential of generic video models in AD.

Experiments on three real-world benchmarks validate this simple design. On NAVSIM~\citep{dauner2024navsim}, SimWAM achieves \textbf{$91.9$} PDMS and offers a favorable trade-off between planning performance and inference latency among current world-model-based planners, as shown in Fig.~\ref{fig:intro}. The same NAVSIM-trained model further transfers zero-shot to nuScenes~\citep{caesar2020nuscenes} without fine-tuning, demonstrating cross-domain generalization. On WOD-E2E and PhysicalAI-Autonomous-Vehicles, SimWAM similarly delivers competitive planning accuracy. We hope SimWAM will serve as a strong and practical baseline for efficient world-action modeling in AD.

The advantages of SimWAM arise from three aspects. First, SimWAM transfers the traffic-dynamics prior of a pretrained video expert to the action expert via joint co-training. Second, thanks to the isolated attention mask, the action expert remains independent of future-frame representations, allowing efficient inference without explicitly generating future frames. Third, the decoupled architecture seamlessly accommodates more advanced video generators without modifying the action expert.

\section{Related Work}

\subsection{Vision-Language-Action Models for Autonomous Driving}
End-to-end autonomous driving integrates perception, prediction, and planning in a unified framework, as demonstrated by UniAD~\citep{hu2023planning} and VAD~\citep{jiang2023vad}, but remains largely dependent on expert trajectory supervision with limited semantic reasoning. Vision-Language-Action (VLA) models~\citep{chen2026recogdrive,fu2025orion,zhou2025hermes,liu2026drivepi} leverage pretrained vision-language representations for semantic reasoning, with AutoVLA~\citep{zhou2026autovla} and ORION~\citep{fu2025orion} further combining reasoning with trajectory generation. FutureSightDrive~\citep{zeng2026futuresightdrive} and ExploreVLA~\citep{sheng2026explorevla} introduce future image generation for scene evolution modeling. In contrast, SimWAM transfers motion priors from a pretrained video generator to a lightweight action expert for direct trajectory prediction.
  
\subsection{World-Action Models for Autonomous Driving}

World-Action Models~\citep{assran2025vjepa2,wang2025adawm,agarwal2025cosmos} jointly learn action prediction and image generation to capture object motion, physical interactions and future scene evolution, as demonstrated in robotics by DreamZero~\citep{ye2026world} and LingBot-VA~\citep{li2026causal}. In autonomous driving, earlier world models such as DriveDreamer~\citep{wang2024drivedreamer} and HERMES~\citep{zhou2025hermes} mainly predicted or generated future driving scenes, while more recent studies~\citep{li2026drivevla,shi2026drivewam} integrate visual world modeling with trajectory planning. Epona~\citep{zhang2025epona} jointly predicts future videos and trajectories via autoregressive diffusion, whereas DriveLaW~\citep{xia2026drivelaw} conditions a diffusion planner on latents produced by a video generator. They couple planning with future visual generation at inference. SimWAM instead uses future-video generation only as training supervision and predicts trajectories directly.

\subsection{Reinforcement Learning for Autonomous Driving}
Imitation learning trains autonomous driving policies to reproduce expert trajectories, but this objective confines learning to demonstrated behavior and indirectly reflects overall driving quality. Reinforcement learning provides a complementary refinement that directly optimizes driving policies with task-level rewards. CarPlanner~\citep{zhang2025carplanner} uses expert-guided rewards to improve large-scale trajectory planning. Raw2Drive~\citep{yang2026raw2drive} refines driving policies with raw sensor inputs and privileged world models. Recent studies~\citep{zhou2026autovla,chen2026recogdrive,fu2025minddrive,sheng2026explorevla} have further introduced reinforcement learning into Vision-Language-Action driving models. MindDrive~\citep{fu2025minddrive} applies online reinforcement learning to refine high-level decisions using feedback from their induced driving trajectories. CritiqueDriveVLM~\citep{liu2026critiquedrivevlm} refines driving reasoning with a verifier and distills this capability into an efficient reasoning model. SimWAM instead applies reward optimization directly to its continuous trajectory-generating action expert.

\section{Preliminary}
\label{sec:prelim}

\textbf{Flow matching.} Rectified flow~\citep{lipman2022flow,liu2022flow} interpolates a clean target $x$ with Gaussian noise $\epsilon\sim\mathcal{N}(0,I)$ as $x_\tau=(1{-}\tau)\,x+\tau\,\epsilon$ ($\tau\in[0,1]$). The network $v_\theta$ learns the target velocity $\epsilon-x$ under conditioning $c$:
\begin{equation}
\mathcal{L}_{\text{FM}}=\mathbb{E}_{x,\epsilon,\tau}\big[\,\|v_\theta(x_\tau,\tau,c)-(\epsilon-x)\|_2^2\,\big].
    \label{eq:fm}
\end{equation}
Sampling integrates the probability-flow ODE $\mathrm{d}x_\tau=v_\theta(x_\tau,\tau,c)\,\mathrm{d}\tau$ from noise ($\tau{=}1$) to data ($\tau{=}0$), so that each noise sample is transported along a deterministic trajectory to one generated sample.

\textbf{From ODE to SDE.} The deterministic ODE generates a single sample per noise input and lacks a tractable transition density, which restricts exploration and precludes policy-gradient optimization. To introduce stochasticity into this deterministic framework, SDE-based GRPO methods~\citep{liu2026flow} adopt sampling with the same marginal distributions as the original ODE:
\begin{equation}
    \mathrm{d}x_\tau=\Big[v_\theta(x_\tau,\tau)+\tfrac{\sigma_\tau^2}{2\tau}\big(x_\tau+(1{-}\tau)\,v_\theta(x_\tau,\tau)\big)\Big]\mathrm{d}\tau+\sigma_\tau\,\mathrm{d}w,\qquad \sigma_\tau=\eta\sqrt{\tfrac{\tau}{1{-}\tau}},
    \label{eq:sde}
\end{equation}
where $\mathrm{d}w$ is a Wiener increment and $\eta$ controls the noise scale. Each Euler-Maruyama step of size $\Delta\tau$ yields an isotropic Gaussian transition $\pi_\theta(x_{\tau-\Delta \tau}\mid x_\tau)=\mathcal{N}\big(\mu_\theta(x_\tau,\tau),\,\sigma_\tau^2\Delta \tau\,I\big)$ with tractable log-likelihoods for importance sampling, in which $\mu_\theta(x_\tau,\tau)$ is the drift-induced mean of the update and reduces to the deterministic ODE step as $\eta$ vanishes.

\section{Method}
\label{sec:method}

\begin{figure}[t]
  \centering
  \includegraphics[width=0.98\textwidth]{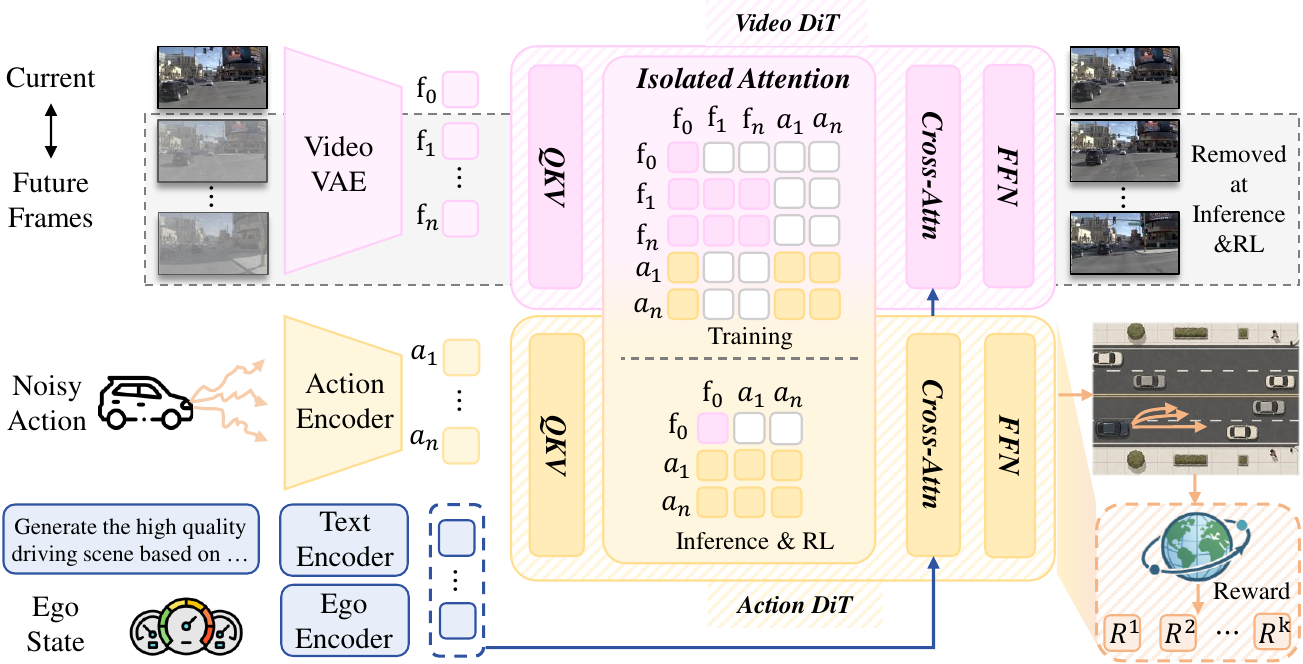}
  \caption{\textbf{Overview of SimWAM.} During training, the video and action DiTs are jointly optimized for future-frame generation and trajectory prediction via shared attention, while the isolated mask prevents the action tokens from accessing future-frame tokens. During inference and reinforcement learning, the model directly predicts trajectories without explicitly generating future frames.}
  \label{fig:model_architecture}
  \vspace{-10pt}
\end{figure}

We present SimWAM as a strong and practical baseline for end-to-end autonomous driving rather than a pursuit of architectural innovation. As illustrated in Fig.~\ref{fig:model_architecture}, the design stays deliberately plain, with a pretrained video expert and a lightweight action expert co-trained by joint flow matching under an isolated attention mask that keeps trajectory prediction free of future frames at inference.

\subsection{Model Architecture}
\label{sec:arch}

\textbf{Problem formulation.} We consider end-to-end trajectory planning from a front-camera observation $o_t$, the ego state $s_t$ containing velocity, acceleration, and yaw rate, and a navigation command $l$. The planner predicts an ego trajectory $a_{t+1:t+H}=(a_{t+1},\ldots,a_{t+H})$ in the ego-vehicle coordinate frame, where $H$ is the number of waypoints and each waypoint $a_i=(x_i,y_i,\psi_i)$ specifies the planned position and heading. Existing driving WAMs~\citep{xia2026drivelaw,shi2026drivewam} commonly follow an imagine-then-act paradigm, factorizing the policy as:
\begin{equation}
    p_\theta(a_{t+1:t+H},\,z_{t+1:t+N}\mid o_t,s_t,l)=p_\theta(z_{t+1:t+N}\mid o_t, s_t, l)\,p_\theta(a_{t+1:t+H}\mid o_t,s_t,l,z_{t+1:t+N}),
    \label{eq:imagine}
\end{equation}
which conditions trajectory generation on the predicted latents $z_{t+1:t+N}$ of $N$ future frames, adding video-generation overhead to planning. Our SimWAM departs from this paradigm by absorbing the traffic-dynamics prior from future-video supervision during co-training, while retaining a plain and direct policy interface without future-scene generation at inference, given by:
\begin{equation}
    p_\theta(a_{t+1:t+H}\mid o_t,s_t,l)=p_\theta\big(a_{t+1:t+H}\mid z(o_t),s_t,l\big),
    \label{eq:direct}
\end{equation}
where $z(o_t)$ is the representation produced from the current observation.

\textbf{Video expert.} We initialize the video Diffusion Transformer (DiT)~\citep{peebles2023scalable} from Wan2.2-5B~\citep{wan2025wan} and use its accompanying video VAE~\citep{kingma2013auto} and T5 text encoder~\citep{raffel2020exploring}. The current frame serves as the clean condition, and the navigation command $l$ is first encoded by the T5 text encoder and then injected into the DiT via cross-attention, while the $N$ future frames are noised and reconstructed with flow matching. This video-generation objective endows the action expert with a traffic-dynamics-aware representation by forcing the model to anticipate how traffic evolves.

\textbf{Action expert.} The action expert is a lightweight Diffusion Transformer with hidden size $d_a{=}1024$. Conditioned on $c=\{z(o_t),s_t,l\}$ with the ego state $s_t$ embedded by a small MLP, it predicts the trajectory velocity field \(v_{\theta}(a^\tau_{t+1:t+H},\tau,c)\) via flow matching. Integrating this velocity field along the probability-flow ODE from Gaussian noise at $\tau{=}1$ to $\tau{=}0$ then yields the planned trajectory through the same flow-matching interface as the video expert.

\textbf{Isolated attention mask.}
\label{sec:mask}
To exploit future-video supervision during training without paying for future-frame synthesis at inference, we adopt the isolated attention mask of Fig.~\ref{fig:model_architecture}. Specifically, the future-frame tokens $z_{t+1:t+N}$ and the action tokens both attend to the current observation $z(o_t)$ while remaining mutually invisible. Consequently, the action expert uses current-frame visual representations $z(o_t)$, together with the ego state and navigation command, without accessing future-frame tokens. Moreover, the same decoupling later allows reinforcement learning to optimize trajectories without any future-frame rollout (\S\ref{sec:rl}).

\textbf{Co-training.} The two experts interact only through shared attention and retain their original architectures. Joint flow matching over video and trajectory modalities allows future-scene prediction to shape the observation representation used for planning. The joint objective is defined as:
\begin{equation}
    \mathcal{L}=\mathcal{L}_{\text{FM}}^{\text{act}}+\lambda\,\mathcal{L}_{\text{FM}}^{\text{vid}},
    \label{eq:joint}
\end{equation}
where $\mathcal{L}_{\text{FM}}^{\text{act}}$ and $\mathcal{L}_{\text{FM}}^{\text{vid}}$ instantiate Eq.~\ref{eq:fm} on the action trajectory $a_{t+1:t+H}$ and the future-frame latents $z_{t+1:t+N}$, and $\lambda$ balances the two terms.

\subsection{Reinforcement Learning}
\label{sec:rl}

Imitation learning fits logged trajectories without directly optimizing the specific driving reward. We therefore refine the action expert with reinforcement learning on a task-level driving reward. As in Sec.~\ref{sec:prelim}, we use the SDE in Eq.~\ref{eq:sde} for stochastic rollouts. Its Gaussian transitions provide exploration and tractable per-step conditional log-likelihoods for policy optimization. To keep the method simple, we adapt the SDE to the trajectory output space with minimal stochasticity. Empirically, SDE integration suits this space better in the later denoising steps than in the earlier ones. We therefore activate it only on the last three steps $\mathcal{W}=\{K-3,K-2,K-1\}$ of the $K$-step chain, whose updates are given by:
\begin{equation}
    a^{(k+1)}=\mu_{\theta,k}+\sigma_k\,\xi_k,\quad
    \sigma_k=\eta\sqrt{\frac{\tau_k\,(\tau_k-\tau_{k+1})}{1-\tau_k}},\quad
    \xi_k\sim\mathcal{N}(0,I),\quad k\in\mathcal{W},
    \label{eq:window}
\end{equation}
where $\mu_{\theta,k}$ is the SDE mean at step $k$ under schedule $\tau_k$. For each scenario, we score $G$ candidates and obtain rewards $r_i$. We compute the group-relative advantage $\hat{A}_i=(r_i-\bar{r})/(\sigma_r+\zeta)$, where $\bar{r}$ and $\sigma_r$ are the group mean and standard deviation and $\zeta$ is a small constant avoiding division by zero. The clipped surrogate objective~\citep{shao2024deepseekmath,guo2025deepseek} is as follows:

\begin{equation}
\begin{gathered}
    \scalebox{0.9}{$\displaystyle
        \mathcal{L}_{\text{RL}}
        =-\mathbb{E}\!\left[
        \min\!\left(
            \tfrac{\pi_\theta(a_i^{(k+1)}\mid a_i^{(k)},c)}
                  {\pi_{\mathrm{old}}(a_i^{(k+1)}\mid a_i^{(k)},c)}
            \hat{A}_i,\ 
            \mathrm{clip}\!\left(
                \tfrac{\pi_\theta(a_i^{(k+1)}\mid a_i^{(k)},c)}
                      {\pi_{\mathrm{old}}(a_i^{(k+1)}\mid a_i^{(k)},c)},
                1{-}\delta,1{+}\delta
            \right)\hat{A}_i
        \right)\right]
        +\beta\mathcal{L}_{\mathrm{anchor}},
    $}\\
    \mathcal{L}_{\mathrm{anchor}}
    =\mathbb{E}_{i,k}\!\left[
        \frac{1}{H\sum_d w_d}
        \sum_{h=1}^{H}\sum_{d=1}^{3}
        w_d\big(\mu_{\theta,ikhd}-\mu_{\mathrm{ref},ikhd}\big)^2
    \right],
\end{gathered}
\label{eq:rl}
\end{equation}

where the expectations run over candidates $i$ and steps $k\in\mathcal{W}$, and $\delta$ is the clip range. The anchor regularizes the SDE mean toward the frozen imitation policy at identical states, with coordinate weights $w=(0.2,1,1)$ leaving the longitudinal coordinate freer to adapt.

\subsection{Flexibility of SimWAM}

The structural simplicity of SimWAM naturally yields flexibility in architecture and model scale. The two experts are parameterized independently and exchange information solely through the attention stream, so their architectures and capacities can be adjusted separately within this interface.

\textbf{Video expert flexibility.} Thanks to this simple interface, SimWAM can seamlessly accommodate different pretrained video experts. The action expert operates on current-frame representations supplied by the video expert rather than future-frame tokens or decoded predictions. Future-video supervision shapes these representations during co-training, while the video expert continues to encode the current observation at inference without generating future frames. Replacing the video backbone with a newer or more driving domain-relevant model thus requires no redesign of the action expert or trajectory objective. In this sense, SimWAM can readily benefit from advances in video generation while preserving the same video-action interface.

\textbf{Scale flexibility.} Within the shared attention interface, the action expert can be scaled independently of the video expert. The video expert supplies current-frame visual features shaped by video pretraining and co-training, while the width of the action expert controls the trade-off between planning capacity and computational cost. This design accommodates different computational budgets without changing the video expert or the training objective.

\section{Experiments}

\subsection{Experimental Setup}
\label{exp: expsetup}

\paragraph{Datasets and benchmarks.}
We evaluate SimWAM on three real-world driving benchmarks. NAVSIM~\citep{dauner2024navsim} provides non-reactive pseudo-simulation with the PDMS metric. The Waymo Open Dataset end-to-end benchmark (WOD-E2E)~\citep{xu2026wod} focuses on long-tail scenarios and is scored by the Rater Feedback Score (RFS) and the average displacement error (ADE). PhysicalAI-Autonomous-Vehicles~\citep{wang2025alpamayo} (PhysicalAI-AV) offers large-scale real-world driving logs evaluated with the ADE and Final Displacement Error (FDE).

\paragraph{Implementation details.}
The video expert is initialized from Wan2.2-5B~\citep{wan2025wan} and the action expert is a lightweight $1024$-wide DiT. For imitation learning (IL), we train for $100$ epochs on NAVSIM, $30$ epochs on WOD-E2E, and $15$ epochs on PhysicalAI-AV. We use front-camera images at resolutions of \(384\times672\) on NAVSIM and \(512\times480\) on WOD-E2E. We subsequently apply reinforcement learning (RL) on NAVSIM and WOD-E2E, while PhysicalAI-AV uses imitation learning only. RL optimizes only LoRA adapters on the action expert, with the SDE restricted to the last three steps. See more details in Appendix~\ref{app:details}.

\subsection{Main Results}

\begin{table}[t!]
  \caption{Comparison with state-of-the-art planners on the NAVSIM \texttt{navtest} benchmark. C denotes camera and L denotes LiDAR.}
  \label{tab:main}
  \centering
  \setlength\tabcolsep{1.8mm}
    \renewcommand{\arraystretch}{1.0}
  \scriptsize
  \begin{tabular}{l l c c c c c c P}
    \toprule
    Method & Reference & Sensors & NC$\uparrow$ & DAC$\uparrow$ & EP$\uparrow$ & TTC$\uparrow$ & C$\uparrow$ & PDMS$\uparrow$ \\
    \midrule
    Human Agent & - & - & 100.0 & 100.0 & 87.5 & 100.0 & 99.9 & 94.8 \\
    \midrule
    \multicolumn{9}{l}{\textit{Traditional E2E planners}} \\
    UniAD~\citep{hu2023planning} & CVPR'23 & 6$\times$C & 97.8 & 91.9 & 78.8 & 92.9 & 100.0 & 83.4 \\
    TransFuser~\citep{chitta2022transfuser} & TPAMI'22 & 3$\times$C+L & 97.7 & 92.8 & 79.2 & 92.8 & 100.0 & 84.0 \\
    DiffusionDrive~\citep{liao2025diffusiondrive} & CVPR'25 & 3$\times$C+L & 98.2 & 96.2 & 82.2 & 94.7 & 100.0 & 88.1 \\
    WoTE~\citep{li2025end} & ICCV'25 & 3$\times$C+L & 98.5 & 96.8 & 81.9 & 94.9 & 99.9 & 88.3 \\
    SeerDrive~\citep{zhang2026future} & NeurIPS'25 & 3$\times$C+L & 98.4 & 97.0 & 83.2 & 94.9 & 99.9 & 88.9 \\
    \midrule
    \multicolumn{9}{l}{\textit{VLM-based planners}} \\
     AutoVLA~\citep{zhou2026autovla} & NeurIPS'25 & 3$\times$C & 98.4 & 95.6 & 81.9 & 98.0 & 99.9 & 89.1 \\
    ReCogDrive~\citep{chen2026recogdrive} & ICLR'26 & 1$\times$C &97.9 &97.3 &\textbf{87.3} &94.9 &100.0  &90.8\\
    DriveVLA-W0 (Flow-Matching)~\citep{li2026drivevla} & ICLR'26 & 1$\times$C & 98.4 & 95.3 &80.9 &95.2 &100.0 & 87.2\\
    DriveVLA-W0 (Query-Base)~\citep{li2026drivevla} & ICLR'26 & 1$\times$C & 98.7 & \textbf{99.1} & 83.3 & 95.3 & 99.3  &90.2 \\
    ExploreVLA~\citep{sheng2026explorevla} & ECCV'26 & 1$\times$C & 98.8 & 98.4 & 83.5 & 96.5 & 99.9 & 90.4 \\
    SGDrive~\citep{li2026sgdrive} & CVPR'26 & 1$\times$C & 98.6 & 97.8 & 85.8 & 96.2 & 100.0 & 91.1 \\
    \midrule
    \multicolumn{9}{l}{\textit{World-model-based planners}} \\
    Epona~\citep{zhang2025epona} & ICCV'25 & 1$\times$C & 97.9 & 95.1 & 80.4 & 93.8 & 99.9 & 86.2 \\
    PWM~\citep{zhao2026forecasting} & NeurIPS'25 & 1$\times$C & 98.6 & 95.9 & 81.8 & 95.4 & 100.0 & 88.1 \\
    DriveLaW~\citep{xia2026drivelaw} & CVPR'26 & 1$\times$C & 99.0 & 97.1 & 81.3 & 96.7 & 100.0 & 89.1 \\
    DriveVA~\citep{liu2026driveva} & ECCV'26 & 1$\times$C & \textbf{99.2} & 97.5&  83.5 &\textbf{98.7} &100.0 & 90.9 \\
    DriveWAM~\citep{shi2026drivewam} & arXiv'26 & 1$\times$C & 98.3 & 98.1 & 84.3 & 95.2 & 100.0 & 90.1 \\
    \midrule
    \rowcolor{defrow} SimWAM (\textbf{ours}) & - & 1$\times$C & 98.5 & 99.1 & 86.5 & 96.1 & 99.8 & \textbf{91.9} \\
    \bottomrule
  \end{tabular}
  \vspace{-8pt}
\end{table}

\begin{table}[t!]
  \caption{Future-video generation quality on \texttt{navtest}. All methods use a 4\,s horizon at 2\,Hz.}
  \label{tab:fvd}
  \centering
  \setlength\tabcolsep{2.7mm}
  \renewcommand{\arraystretch}{1.0}
  \scriptsize
  \begin{tabular}{l ccccccc}
    \toprule
    & SVD & DrivingGPT  & Uni-World VLA & PWM  & ForgeDrive & DriveDreamer-Policy & SimWAM (\textbf{ours}) \\
    \midrule
    FVD$\downarrow$ & 227.5 & 142.6 &141.8 & 86.0 & 69.2 & 53.6 & \textbf{36.3} \\
    \bottomrule
  \end{tabular}
\end{table}

\begin{table}[t!]
  \caption{Open-loop planning on the WOD-E2E benchmark. ADE denotes the average displacement error, and RFS the Rater Feedback Score. $*$ indicates reinforcement learning.}
  \label{tab:waymo}
  \centering
  \small
  \vspace{-6pt}
  \begin{minipage}[t]{0.49\linewidth}
    \centering
    (a) Validation split.
    \setlength{\tabcolsep}{1.2mm}
    \renewcommand{\arraystretch}{1.0}
    \scriptsize
    \begin{tabular}{l c c c}
      \toprule
      Method & ADE@3s$\downarrow$ & ADE@5s$\downarrow$ & RFS$\uparrow$ \\
      \midrule
      Human Driver & - & - & 8.13 \\
      VAD~\citep{jiang2023vad} & 3.19 & 5.81 & 4.45 \\
      UniAD~\citep{hu2023planning} & 6.50 & 10.81 & 5.78 \\
      RAP-DINO~\citep{feng2026rap} & 0.97 & 2.20 & 7.91 \\
      \midrule
      SimWAM (\textbf{ours}) & 0.98 & 2.29 & 7.99 \\
      \rowcolor{defrow} SimWAM (\textbf{ours})$^*$ & 0.98 & 2.27 & \textbf{8.29} \\
      \bottomrule
    \end{tabular}
  \end{minipage}
  \hfill
  \begin{minipage}[t]{0.49\linewidth}
    \centering
    (b) Test split.
    \setlength{\tabcolsep}{1.1mm}
    \renewcommand{\arraystretch}{0.82}
    \scriptsize
    \begin{tabular}{l c c c}
      \toprule
      Method & ADE@3s$\downarrow$ & ADE@5s$\downarrow$ & RFS$\uparrow$ \\
      \midrule
      UniPlan~\citep{liao2025diffusiondrive} & 1.31 & 2.99 & 7.78 \\
      dVLM-AD~\citep{ma2026dvlmad} & 1.29 & 3.02 & 7.63 \\
      HMVLM~\citep{wang2025hmvlm} & 1.33 & 3.07 & 7.74 \\
      AutoVLA$^*$~\citep{zhou2026autovla} & 1.35 & 2.96 & 7.56 \\
      NoRD$^*$~\citep{rawal2026nord} & 1.25 & - & 7.71 \\
      \midrule
      SimWAM (\textbf{ours}) & 1.22 & 2.69 & 7.77 \\
      \rowcolor{defrow} SimWAM (\textbf{ours})$^*$ & 1.21 & 2.68 & \textbf{7.84} \\
      \bottomrule
    \end{tabular}
  \end{minipage}
  \vspace{-8pt}
\end{table}

\noindent\textbf{NAVSIM.} As shown in Tab.~\ref{tab:main}, our method achieves $91.9$ PDMS on the test split with a single front camera, surpassing recent VLM-based and world-model-based end-to-end planners. It exceeds SGDrive~\citep{li2026sgdrive} by $0.8$ points and ExploreVLA~\citep{sheng2026explorevla}, a VLA with future image generation, by $1.5$ points. Notably, our future-synthesis-free model further bests the imagine-then-act planners DriveLaW~\citep{xia2026drivelaw} and DriveVA~\citep{liu2026driveva} by $2.8$ and $1.0$ points. Moreover, the co-trained video expert achieves the lowest FVD of $36.3$ in Tab.~\ref{tab:fvd}, compared with $53.6$ for the next-best method. These results show that SimWAM combines strong planning performance with high-quality video generation, while keeping trajectory inference independent of future synthesis. More results appear in Appendix~\ref{app:moreexp}.

\noindent\textbf{WOD-E2E.} Tab.~\ref{tab:waymo} compares open-loop planning on the WOD-E2E validation and test splits. Although reinforcement learning is performed on the validation split, the imitation-trained model alone already surpasses the strongest baseline RAP-DINO~\citep{feng2026rap} there with an RFS of $7.99$. On the held-out Waymo test split, the imitation-trained model already surpasses all compared methods in ADE, and reinforcement learning generalizes to this split as well, improving RFS from $7.77$ to $7.84$. Notably, SimWAM achieves this without language reasoning or test-time future synthesis, highlighting the strength of combining joint video co-training with reinforcement learning.

\begin{table}[t!]
  \caption{Comparison on the $1{,}000$-clip test subset of the PhysicalAI-Autonomous-Vehicles benchmark~\citep{shi2026drivewam}. Params. denotes the number of model parameters. $*$ indicates evaluation with the released checkpoint, which supports prediction up to $3$\,s only.}
  \label{tab:physicalai_av}
  \centering
  \setlength\tabcolsep{3.0mm}
  \renewcommand{\arraystretch}{1.0}
  \scriptsize
  \begin{tabular}{l c c c c c c}
    \toprule
    Method & Source & Params. & ADE@3s$\downarrow$ & FDE@3s$\downarrow$ & ADE@4s$\downarrow$ & FDE@4s$\downarrow$ \\
    \midrule
    VaVAM$*$~\citep{bartoccioni2025vavim} & Valeo & 1.3B & 2.31 & 4.32 & - & - \\
    Alpamayo-1.5~\citep{wang2025alpamayo} & NVIDIA & 10B & 0.80 & 2.31 & 1.44 & 4.18 \\
    DriveWAM~\citep{shi2026drivewam} & - & 5B + 8B & 0.47 & 1.35 & 0.83 & 2.47 \\
    \midrule
    \rowcolor{defrow} SimWAM (\textbf{ours}) & - & 6B & 0.40 & 1.08 & 0.69 & 1.96 \\
    \bottomrule
  \end{tabular}
  \vspace{-8pt}
\end{table}

\begin{table}[t!]
  \caption{Zero-shot generalization on the nuScenes open-loop planning benchmark. $*$ represents only using the front camera as input.}
  \label{tab:generalization}
  \centering
  \renewcommand{\arraystretch}{1.0}
  \setlength\tabcolsep{1.4mm}
  \scriptsize
  \begin{tabular}{l c c c c c c c c c c c}
    \toprule
    \multirow{2}{*}{Method} & \multirow{2}{*}{Finetune} & \multirow{2}{*}{Input} & \multirow{2}{*}{Auxiliary Supervision} & \multicolumn{4}{c}{L2 (m)\,$\downarrow$} & \multicolumn{4}{c}{Collision Rate (\%)\,$\downarrow$} \\
    \cmidrule(lr){5-8} \cmidrule(l){9-12}
    & & & & 1\,s & 2\,s & 3\,s & Avg. & 1\,s & 2\,s & 3\,s & Avg. \\
    \midrule
    ST-P3~\citep{hu2022st} & \ding{51} & Camera & Map\&Box\&Depth & 1.33 & 2.11 & 2.90 & 2.11 & 0.23 & 0.62 & 1.27 & 0.71 \\
    UniAD~\citep{hu2023planning} & \ding{51} & Camera & Map\&Box\&Motion & 0.48 & 0.96 & 1.65 & 1.03 & 0.05 & 0.17 & 0.71 & 0.31 \\
    VAD-Base~\citep{jiang2023vad} & \ding{51} & Camera & Map\&Box\&Motion & 0.54 & 1.15 & 1.98 & 1.22 & 0.04 & 0.39 & 1.17 & 0.53 \\
    GenAD~\citep{zheng2024genad} & \ding{51} & Camera & Map\&Box\&Motion & 0.36 & 0.83 & 1.55 & 0.91 & 0.06 & 0.23 & 1.00 & 0.43 \\
    Epona~\citep{zhang2025epona} & \ding{51} & Camera$^{*}$ & None & 0.61 & 1.17 & 1.98 & 1.25 & 0.01 & 0.22 & 0.85 & 0.36 \\
    \midrule
    DriveVA~\citep{liu2026driveva} & \ding{55} & Camera$^{*}$ & None & 0.33 & \textbf{0.76} & \textbf{1.43} & \textbf{0.84} & 0.00 & 0.07 & 0.12 & 0.06 \\
    DriveWAM~\citep{shi2026drivewam} & \ding{55} & Camera$^{*}$ & None & \textbf{0.28} & 0.81 & 1.80 & 0.96 & 0.00 & 0.05 & 0.14 & 0.06 \\
    \rowcolor{defrow} SimWAM (\textbf{ours}) & \ding{55} & Camera$^{*}$ & None & 0.29 & 0.82 & 1.77 & 0.96 & 0.00 & \textbf{0.03} & \textbf{0.11} & \textbf{0.05} \\
    \bottomrule
  \end{tabular}
  \vspace{-10pt}
\end{table}

\noindent\textbf{PhysicalAI-AV.} We further examine open-loop trajectory prediction on PhysicalAI-Autonomous-Vehicles in Tab.~\ref{tab:physicalai_av}, which reports only the imitation-trained model. Trained on only $65$K samples, SimWAM attains the lowest ADE and FDE at both the $3$s and $4$s horizons, surpassing VaVAM~\citep{bartoccioni2025vavim} and Alpamayo-1.5~\citep{wang2025alpamayo} despite their far larger training corpora. More notably, DriveWAM~\citep{shi2026drivewam} builds on the same video expert but relies on test-time future-frame synthesis. SimWAM outperforms it without any explicit future-video generation.

\noindent\textbf{Zero-shot generalization on nuScenes.} We further evaluate the NAVSIM-trained SimWAM on the nuScenes~\citep{caesar2020nuscenes} open-loop benchmark without fine-tuning. As shown in Tab.~\ref{tab:generalization}, despite using neither nuScenes supervision nor auxiliary annotations, SimWAM achieves an average L2 error and collision rate comparable to or better than those of strong zero-shot baselines, indicating that its learned dynamics-aware representation transfers beyond the training benchmark.

Overall, despite being deliberately plain and forgoing bells and whistles such as language reasoning or test-time future synthesis, SimWAM consistently matches or even surpasses substantially more complex planners across diverse benchmarks, clearly demonstrating that a simple world-action design is already remarkably robust and generalizable for end-to-end driving.

\subsection{Ablation Studies And Analysis}

Unless otherwise noted, all ablations report the imitation-trained world-action model on the NAVSIM \texttt{navtest} split, and all inference latencies are measured on a single NVIDIA A100 GPU.

\noindent\textbf{Analysis on the attention mask.} The attention pattern determines how information flows between the two experts, and we compare three alternatives in Tab.~\ref{tab:ablation_mask}. Both bidirectional and action$\to$video attention tie action prediction to future video tokens, forcing the model to instantiate future-frame representations at inference. In contrast, our isolated mask lets the action expert attend only to the current observation, so trajectory inference never touches future-frame synthesis. Despite its simpler dependency structure, the isolated mask achieves the best PDMS of $90.3$ while predicting the trajectory nearly $3\times$ faster than the masks that condition on future frames. These results suggest that future-video prediction can serve as training supervision rather than a test-time requirement for effective planning.

\begin{figure}[t!]
  \centering
  \begin{minipage}[t]{0.48\linewidth}
    \centering
    \includegraphics[width=\linewidth]{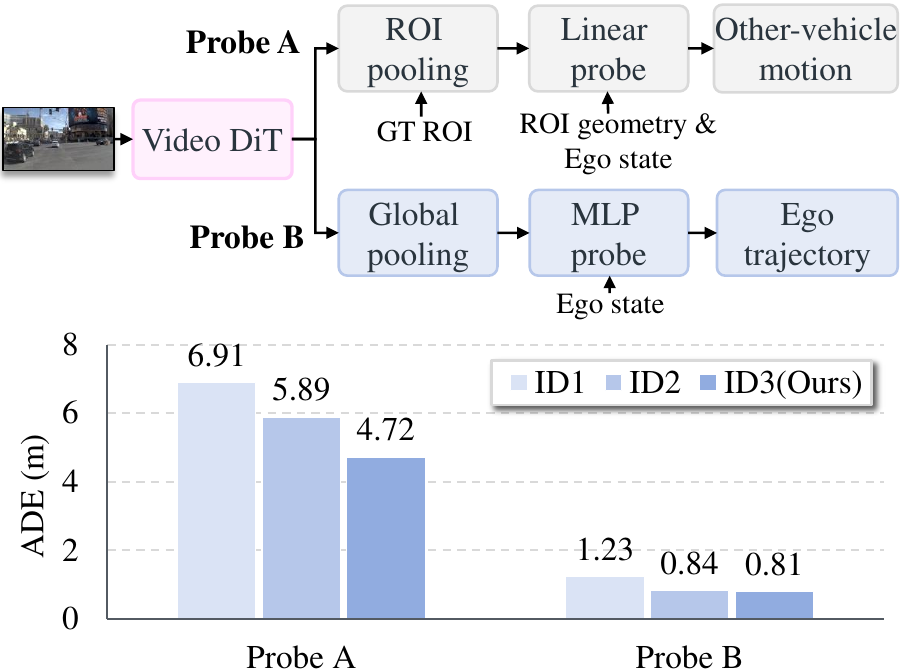}
    \caption{Probe-based representation analysis on the current-frame features of the video expert.}
    \label{fig:ablation_probe}
  \end{minipage}
  \hfill
  \begin{minipage}[t]{0.48\linewidth}
    \centering
    \includegraphics[width=\linewidth]{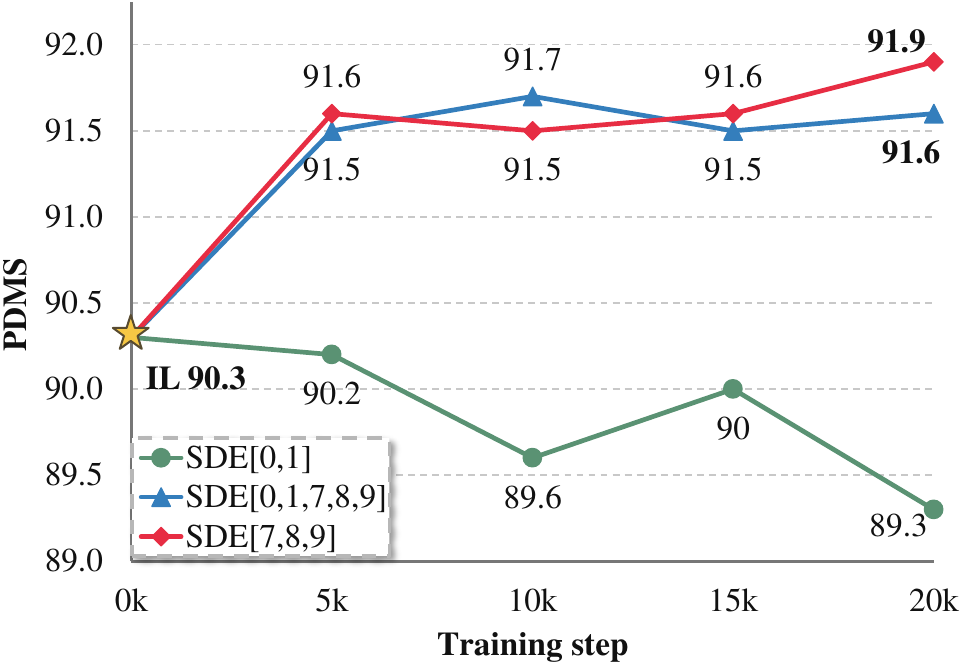}
    \caption{Ablation on the placement of the SDE steps during RL.}
    \label{fig:ablation_sde}
  \end{minipage}
  \vspace{-8pt}
\end{figure}

\begin{table}[t!]
  \centering
  \footnotesize
  \begin{minipage}[t]{0.44\linewidth}
    \centering
    \setlength{\tabcolsep}{0.5mm}
    \renewcommand{\arraystretch}{1.1}
    \captionof{table}{Ablation on attention mask.}
    \label{tab:ablation_mask}
    \scriptsize
    \vspace{-4pt}
    \begin{tabular}{lcccc Q c}
      \toprule
      Mask & NC & DAC & EP & TTC & PDMS & Latency (ms)$\downarrow$ \\
      \midrule
      Bidirectional & 98.4 & \textbf{98.0} & \textbf{84.7} & 95.1 & 90.2 & 1512 \\
      Action$\to$video & 98.5 & 97.8 & 84.3 & 95.5 & 90.1 & 1499 \\
      \rowcolor{defrow} Isolated & \textbf{98.6} & 98.0 & 84.0 & \textbf{95.8} & \textbf{90.3} & \textbf{518} \\
      \bottomrule
    \end{tabular}
  \end{minipage}
  \hfill
  \begin{minipage}[t]{0.54\linewidth}
    \centering
    \setlength{\tabcolsep}{0.75mm}
    \renewcommand{\arraystretch}{0.88}
    \captionof{table}{Video expert flexibility. RL in parentheses.}
    \scriptsize
    \label{tab:ablation_video}
    \vspace{-4pt}
    \begin{tabular}{l ccccc}
      \toprule
      Video model & NC & DAC & EP & TTC & PDMS \\
      \midrule
      LTX-Video & 98.2\,{\tiny(98.6)} & 97.2\,{\tiny(98.6)} & 83.0\,{\tiny(85.5)} & 94.4\,{\tiny(95.3)} & 88.7\,{\tiny(90.9)} \\
      Wan2.1-1.3B & 98.6\,{\tiny(98.6)} & \textbf{98.1}\,{\tiny(99.1)} & 84.0\,{\tiny(86.1)} & 95.9\,{\tiny(96.3)} & 90.2\,{\tiny(91.8)} \\
      Cosmos2.5 & \textbf{98.7}\,{\tiny(98.8)} & 98.0\,{\tiny(99.2)} & \textbf{84.2}\,{\tiny(87.0)} & \textbf{95.9}\,{\tiny(96.4)} & \textbf{90.4}\,{\tiny(92.4)} \\
      \rowcolor{defrow} Wan2.2-5B & 98.6\,{\tiny(98.5)} & 98.0\,{\tiny(99.1)} & 84.0\,{\tiny(86.5)} & 95.8\,{\tiny(96.1)} & 90.3\,{\tiny(91.9)} \\
      \bottomrule
    \end{tabular}
  \end{minipage}
\vspace{-10pt}
\end{table}

\begin{table}[t!]
  \centering
  \footnotesize
  \begin{minipage}[t]{0.44\linewidth}
    \centering
    \setlength{\tabcolsep}{0.9mm}
  \renewcommand{\arraystretch}{1.18}
  \captionof{table}{Ablation on action expert scaling.}
  \scriptsize
  \label{tab:ablation_action}
   \vspace{-4pt}
  \begin{tabular}{lccccc c}
    \toprule
    \makecell[c]{Action\\Params (B)} & NC & DAC & EP & TTC & PDMS & FLOPs (G)$\downarrow$\\
    \midrule
    $0.21$  & 98.6 & 97.8 & 84.0 & 95.4 & 89.9 & 27.3 \\
    $0.45$ & 98.6 & 97.9 & 83.8 & \textbf{95.9} & 90.1 & 55.3 \\
    \rowcolor{defrow} $1.02$ & \textbf{98.6} & \textbf{98.0} & \textbf{84.0} & 95.8 & \textbf{90.3} & 112.8 \\
    \bottomrule
  \end{tabular}
  \end{minipage}
  \hfill
  \begin{minipage}[t]{0.54\linewidth}
    \centering
    \setlength{\tabcolsep}{0.8mm}
    \renewcommand{\arraystretch}{0.88}
    \captionof{table}{Ablation on the co-training signals.}
    \label{tab:ablation_comp}
    \scriptsize
    \vspace{-4pt}
    \begin{tabular}{c ccc ccccc}
      \toprule
      ID & \makecell[c]{Video Expert\\Training} & \makecell[c]{Future-Video\\Supervision} & RL & NC & DAC & EP & TTC & PDMS \\
      \midrule
      1 & - & - & - & 98.0 & 95.2 & 81.0 & 93.8 & 86.5 \\
      2 & \ding{51} & - & - & 98.2 & 96.8 & 82.6 & 94.5 & 88.4 \\
      3 & \ding{51} & \ding{51} & - & \textbf{98.6} & 98.0 & 84.0 & 95.8 & 90.3 \\
      \rowcolor{defrow} 4 & \ding{51} & \ding{51} & \ding{51} & 98.5 & \textbf{99.1} & \textbf{86.5} & \textbf{96.1} & \textbf{91.9} \\
      \bottomrule
    \end{tabular}
  \end{minipage}
\vspace{-12pt}
\end{table}

\noindent\textbf{Analysis on video expert flexibility.}
SimWAM accommodates diverse pretrained video generators through a unified attention interface, as summarized in Tab.~\ref{tab:ablation_video}. Wan2.1-1.3B and Wan2.2-5B attain comparable PDMS of $90.2$ and $90.3$ under IL and $91.8$ and $91.9$ after RL, confirming that our method remains independent of any particular video backbone. Notably, the newer Cosmos-Predict2.5~\citep{ali2025world}, pretrained on driving videos, provides a stronger driving-dynamics prior and achieves the best PDMS of $90.4$ under IL and $92.4$ after RL. By comparison, the lightweight LTX-Video~\citep{hacohen2024ltx} reaches $88.7$ under IL and $90.9$ after RL, indicating that the quality of this driving prior still matters. These results highlight that SimWAM can seamlessly absorb stronger and more driving-relevant priors from advanced video generators.

\noindent\textbf{Analysis on action expert scalability.} The parameter-decoupled two-expert design further allows independent scaling of the action expert, as listed in Tab.~\ref{tab:ablation_action}. Under imitation learning, PDMS climbs from $89.9$ to $90.3$ as the action DiT grows from $0.21$B to $1.02$B with a fourfold FLOPs increase, and this scaling preserves the video expert and training objective. The width can therefore be traded against the deployment budget. Unless otherwise stated, we use the $1.02$B action expert because it achieves the highest PDMS.

\noindent\textbf{Analysis on the co-trained representation.} Tab.~\ref{tab:ablation_comp} and Fig.~\ref{fig:ablation_probe} jointly examine how co-training shapes the representation that drives planning. Fine-tuning the video expert using the action loss improves PDMS from $86.5$ (ID 1) to $88.4$ (ID 2). Adding future-video supervision further increases PDMS to $90.3$ (ID 3). This consistent progression indicates that supervision over future frames instills a strong traffic-dynamics prior that directly benefits trajectory planning. Action RL (ID 4) then contributes a further $1.6$, demonstrating that optimizing against the driving reward mitigates the conservative predictions of the imitation-trained policy.

Beyond planning accuracy, we verify this prior at the representation level. Since SimWAM reads only current-frame features when planning at inference, we probe precisely these features of the video DiT as trained under each configuration above. Two lightweight probes are attached to them. Probe A linearly predicts the motion of surrounding vehicles from ROI-pooled features, whereas Probe B regresses the ego trajectory from globally pooled ones. Both are trained on a $10$k-scene subset and evaluated on the test split. Under both probes, the co-trained features (ID 3) attain a markedly lower ADE than those of IDs 1 and 2, indicating that co-training distills future scene dynamics into the current-frame representation and yields features substantially more pertinent to planning.

\noindent\textbf{Ablation on the placement of the SDE steps.}
Fig.~\ref{fig:ablation_sde} compares where the SDE noise enters the denoising chain. Applying it only to the first two steps lowers PDMS from the imitation baseline of $90.3$ to $89.3$, whereas restricting it to the last three steps achieves the highest final PDMS of $91.9$. Adding early-step noise to this setting offers no consistent improvement, with a final PDMS of $91.6$. We argue that subsequent deterministic steps can attenuate early perturbations, thereby limiting trajectory diversity and reward differences. Late-step noise may instead influence the final trajectories more directly, enabling more effective exploration and more informative reward signals. We therefore confine the SDE to the last three steps.

\subsection{Qualitative Results}

\begin{figure}[t!]
  \centering
\includegraphics[width=0.98\textwidth]{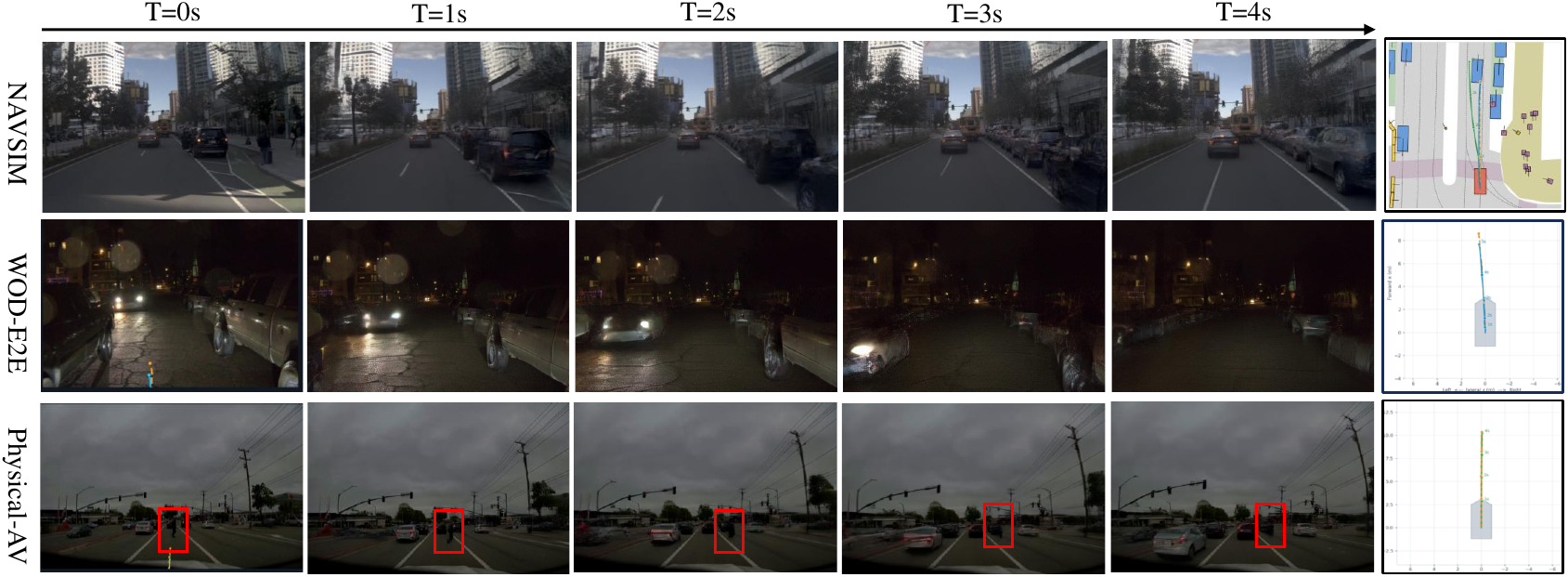}
  \caption{Future-video and trajectory predictions of SimWAM on NAVSIM, WOD-E2E, and PhysicalAI-AV. The red box marks a pedestrian entering the crosswalk.}
  \label{fig:qualitative}
\vspace{-10pt}
\end{figure}

As shown in Fig.~\ref{fig:qualitative}, we visualize the predicted future frames and the planned trajectories of SimWAM on NAVSIM, WOD-E2E, and PhysicalAI-Autonomous-Vehicles. Remarkably, across all these scenes the model delivers accurate future predictions with feasible trajectories, whether in narrow urban streets, night-time glare, or signalized intersections. Most notably, when a pedestrian enters the crosswalk, SimWAM plans a smooth deceleration to yield to the pedestrian, while the predicted frames anticipate the pedestrian’s motion. The generated videos further remain largely temporally coherent and consistent with the planned trajectories.

\section{Conclusion}

In this paper, we present \textbf{SimWAM}, a simple yet flexible world-action model that treats future-video prediction purely as a training-time supervision signal. Through joint flow matching under an isolated attention mask, the lightweight action expert absorbs the traffic-dynamics prior of a pretrained video expert while remaining free of future frames at inference. The decoupled design further allows the video backbone to be replaced and the action expert to be scaled independently, while reinforcement learning aligns trajectory generation with the driving reward beyond imitation. Empirically, SimWAM achieves $91.9$ PDMS on NAVSIM with a favorable trade-off between accuracy and latency among world-model-based planners, transfers zero-shot to nuScenes, and similarly attains competitive planning accuracy on WOD-E2E and PhysicalAI-Autonomous-Vehicles. These results suggest that training-time world modeling offers a practical route to efficient and generalizable end-to-end driving.

\bibliography{refs}
\bibliographystyle{iclr2027_conference}

\clearpage
\appendix

\section{Appendix Overview}
The appendix is organized as follows:
\begin{itemize}
  \setlength\itemsep{0pt}
  \item[--] Appendix B: Dataset statistics, evaluation metrics, and training details.
  \item[--] Appendix C: Additional results on NAVSIM-v2 and further ablations.
  \item[--] Appendix D: More qualitative results.
\end{itemize}

\section{More Details on Datasets and Experimental Setup}
\label{app:details}

\subsection{Datasets}
\label{app:data}

NAVSIM is derived from the OpenScene~\citep{openscene2023} subset of nuPlan~\citep{caesar2021nuplan}. We train on the \texttt{navtrain} split with $85{,}109$ samples and evaluate on the \texttt{navtest} split with $12{,}146$ samples. On WOD-E2E~\citep{xu2026wod} we exclude training samples whose future frames are incomplete, which leaves $333{,}537$ samples, and evaluate on the rater-annotated validation and test splits. PhysicalAI-Autonomous-Vehicles contains roughly $1{,}700$ hours of driving logs organized into $306{,}152$ clips of $20$ seconds, with $153{,}625$ clips for training, $90{,}928$ for validation, and $61{,}599$ for testing. We use the front-view camera stream and ego-motion labels. Following DriveWAM~\citep{shi2026drivewam}, we use $65$K training samples from the training clips for imitation learning and evaluate on the same $1{,}000$-clip test subset for a consistent comparison.

\subsection{Evaluation Metrics}
\label{app:metrics}

\noindent\textbf{NAVSIM PDMS.} NAVSIM~\citep{dauner2024navsim} evaluates each predicted trajectory via non-reactive pseudo-simulation over $4$\,s, scoring no at-fault collision (NC), drivable area compliance (DAC), ego progress (EP), time-to-collision (TTC) within bound, and comfort (C). The scene score is:
\begin{equation}
\mathrm{PDMS} = \mathrm{NC}\cdot\mathrm{DAC}\cdot\frac{5\,\mathrm{EP}+5\,\mathrm{TTC}+2\,\mathrm{C}}{12}.
\end{equation}

\noindent\textbf{NAVSIM-v2 EPDMS and navhard.} NAVSIM-v2~\citep{Cao2025CORL} adds driving direction compliance (DDC), traffic light compliance (TLC), lane keeping (LK), and extended comfort (EC), and refines C into history comfort (HC). Its scene score is:
\begin{equation}
\mathrm{EPDMS} = \mathrm{NC}\cdot\mathrm{DAC}\cdot\mathrm{DDC}\cdot\mathrm{TLC}\cdot\frac{5\,\mathrm{EP}+5\,\mathrm{TTC}+2\,\mathrm{LK}+2\,\mathrm{HC}+2\,\mathrm{EC}}{16},
\end{equation}
where a term scores one whenever its human counterpart scores zero. NAVSIM-v2 further introduces \texttt{navhard}, a safety-critical subset evaluated in two stages. The first stage scores the planner on real-world scenarios, and the second re-evaluates the corresponding synthesized scenarios with traffic agents reacting to the ego vehicle.

\noindent\textbf{WOD-E2E RFS.} WOD-E2E~\citep{xu2026wod} compares a prediction with three human-rated references $p$, each scored $s_p\in[0,10]$. At $t\in\{3,5\}$\,s, let $d_{p,t}$ be the larger axis-wise displacement between the prediction and reference $p$, normalized by the speed-scaled trust-region thresholds. The per-reference and final scores are:
\begin{equation}
\begin{aligned}
q_{p,t}&=s_p\,0.1^{\max(d_{p,t}-1,0)},\qquad S=\frac12\sum_{t\in\{3,5\}}\max_p q_{p,t},\\
\mathrm{RFS}&=\begin{cases}
S,&\text{if } \max_{t\in\{3,5\}}d_{p,t}\le1 \text{ for some } p,\\
\max(S,4),&\text{otherwise}.
\end{cases}
\end{aligned}
\end{equation}
The official test RFS averages scenario-cluster scores, with the lower bound applied after horizon averaging unless both endpoints lie within one reference's trust region. ADE@3s/@5s averages position errors at $4$\,Hz through each horizon.

\subsection{Training Details}
\label{app:training}

On NAVSIM the action expert predicts $8$ waypoints over $4$\,s at $2$\,Hz while the video expert predicts the corresponding $8$ future frames. On WOD-E2E we predict $20$ waypoints over $5$\,s at $4$\,Hz, and on PhysicalAI we report the ADE and FDE over $3$\,s and $4$\,s horizons. On both WOD-E2E and PhysicalAI the video expert predicts $4$ future frames over $4$\,s at $1$\,Hz. All imitation-learning runs adopt AdamW~\citep{loshchilov2019adamw} and a cosine learning rate schedule with an initial learning rate of $10^{-4}$, with $\lambda{=}1$ across the three datasets. During reinforcement learning (RL), we optimize rank-$16$ LoRA adapters~\citep{hu2022lora} with scale $\alpha{=}32$ on the attention projections of the action expert. We sample $G{=}8$ trajectories per scenario with noise level $\eta{=}0.1$ restricted to the last $3$ of the $10$ sampling steps, and use a learning rate of $5{\times}10^{-5}$, an anchor weight of $\beta{=}50$, coordinate weights $w{=}(0.2,1,1)$, and $\zeta{=}10^{-8}$. On NAVSIM, RL trains for $20{,}000$ steps on the \texttt{navtrain} scenes whose imitation PDMS falls below $90$, while evaluation covers the full \texttt{navtest} split. On WOD-E2E, RL optimizes the RFS reward for $10$ epochs on the validation split. Imitation learning runs on $48$ A100 GPUs with total batch size $192$, while RL runs on a single A100 GPU.

\begin{table}[t!]
  \caption{Performance on the NAVSIM-v2 \texttt{navtest} benchmark. $*$ indicates training with reinforcement learning. The best learned result in each column is shown in \textbf{bold}.}
  \label{tab:navtest_v2}
  \centering
  \setlength\tabcolsep{1.5mm}
  \renewcommand{\arraystretch}{0.9}
  \scriptsize
  \begin{tabular}{l c c c c c c c c c c c}
    \toprule
    Method & Reference & NC$\uparrow$ & DAC$\uparrow$ & DDC$\uparrow$ & TLC$\uparrow$ & EP$\uparrow$ & TTC$\uparrow$ & LK$\uparrow$ & HC$\uparrow$ & EC$\uparrow$ & EPDMS$\uparrow$ \\
    \midrule
    Human Agent & - & 100.0 & 100.0 & 99.8 & 100.0 & 87.4 & 100.0 & 100.0 & 98.1 & 90.1 & 90.3 \\
    \midrule
    \multicolumn{12}{l}{\textit{Traditional E2E planners}} \\
    TransFuser~\citep{chitta2022transfuser} & TPAMI'22 & 96.9 & 89.9 & 97.8 & 99.7 & 87.1 & 95.4 & 92.7 & 98.3 & 87.2 & 76.7 \\
    DiffusionDrive~\citep{liao2025diffusiondrive} & CVPR'25 & 98.2 & 95.9 & 99.4 & 99.8 & 87.5 & 97.3 & 96.8 & 98.3 & \textbf{87.7} & 84.5 \\
    \midrule
    \multicolumn{12}{l}{\textit{VLM-based planners}} \\
    ReCogDrive$*$~\citep{chen2026recogdrive} & ICLR'26 & 98.3 & 95.2 & 99.5 & 99.8 & 87.1 & 97.5 & 96.6 & 98.3 & 86.5 & 83.6 \\
    SGDrive~\citep{li2026sgdrive} & CVPR'26 & 98.6 & 94.3 & 99.5 & \textbf{99.9} & 86.0 & 97.9 & 96.1 & 98.3 & 85.9 & 86.2 \\
    \midrule
    \multicolumn{12}{l}{\textit{World-model-based planners}} \\
    DriveVLA-W0~\citep{li2026drivevla} & ICLR'26 & 98.5 & \textbf{99.1} & 98.0 & 99.7 & 86.4 & 98.1 & 93.2 & 97.9 & 58.9 & 86.1 \\
    DriveLaW~\citep{xia2026drivelaw} & CVPR'26 & \textbf{98.7} & 96.9 & 99.6 & 99.8 & 87.5 & 98.3 & 97.6 & \textbf{98.4} & 77.4 & 88.6 \\
    \midrule
    \rowcolor{defrow} SimWAM (\textbf{ours}) & - & 98.6 & 98.0 & \textbf{99.7} & \textbf{99.9} & 87.5 & \textbf{98.4} & \textbf{97.9} & 98.3 & 84.4 & \textbf{90.2} \\
    \bottomrule
  \end{tabular}
  \vspace{-12pt}
\end{table}

\begin{table}[t!]
  \caption{Performance on the NAVSIM-v2 \texttt{navhard} benchmark. S1/S2 denote the per-stage results of the official two-stage protocol, and EPDMS is the overall score. The best learned result in each column is shown in \textbf{bold}.}
  \label{tab:navhard_v2}
  \centering
  \setlength\tabcolsep{1.2mm}
  \renewcommand{\arraystretch}{0.9}
  \scriptsize
  \begin{tabular}{l c c c c c c c c c c c c}
    \toprule
    Method & Reference & Stage & NC$\uparrow$ & DAC$\uparrow$ & DDC$\uparrow$ & TLC$\uparrow$ & EP$\uparrow$ & TTC$\uparrow$ & LK$\uparrow$ & HC$\uparrow$ & EC$\uparrow$ & EPDMS$\uparrow$ \\
    \midrule
    \multicolumn{13}{l}{\textit{Traditional E2E planners}} \\
    \multirow{2}{*}{TransFuser~\citep{chitta2022transfuser}} & \multirow{2}{*}{TPAMI'22} & S1 & 96.2 & 79.5 & 99.1 & 99.5 & 84.1 & 95.1 & 94.2 & 97.5 & 79.1 & \multirow{2}{*}{23.1} \\
    & & S2 & 77.7 & 70.2 & 84.2 & 98.0 & 85.1 & 75.6 & 45.4 & 95.7 & \textbf{75.9} & \\
    \multirow{2}{*}{DiffusionDrive~\citep{liao2025diffusiondrive}} & \multirow{2}{*}{CVPR'25} & S1 & 96.8 & 86.0 & 98.8 & 99.3 & 84.0 & 95.8 & 96.7 & 97.6 & \textbf{79.6} & \multirow{2}{*}{27.5} \\
    & & S2 & 80.1 & 72.8 & 84.4 & 98.4 & 85.9 & 76.6 & 46.4 & 96.3 & 72.8 & \\
    \midrule
    \multicolumn{13}{l}{\textit{VLM-based planners}} \\
    \multirow{2}{*}{SGDrive~\citep{li2026sgdrive}} & \multirow{2}{*}{CVPR'26} & S1 & 95.8 & 87.6 & 97.8 & \textbf{99.8} & 84.4 & 94.7 & 92.9 & \textbf{97.8} & 28.9 & \multirow{2}{*}{25.5} \\
    & & S2 & 79.4 & 65.4 & 79.1 & \textbf{98.9} & 88.9 & 75.3 & 42.7 & 96.4 & 29.6 & \\
    \multirow{2}{*}{ReCogDrive~\citep{chen2026recogdrive}} & \multirow{2}{*}{ICLR'26} & S1 & 96.4 & 78.9 & 98.7 & \textbf{99.8} & 82.6 & 95.6 & 94.4 & 97.6 & 74.2 & \multirow{2}{*}{25.7} \\
    & & S2 & 80.2 & 65.0 & 82.4 & 98.7 & 85.2 & 76.9 & 43.8 & \textbf{96.6} & 71.8 & \\
    \midrule
    \multicolumn{13}{l}{\textit{World-model-based planners}} \\
    \multirow{2}{*}{DriveVLA-W0~\citep{li2026drivevla}} & \multirow{2}{*}{ICLR'26} & S1 & 96.8 & 83.3 & 99.0 & 99.6 & \textbf{84.6} & 95.3 & 96.4 & 97.6 & 78.2 & \multirow{2}{*}{24.4} \\
    & & S2 & 76.8 & 64.3 & 79.9 & 98.3 & \textbf{89.2} & 75.0 & 46.8 & 95.8 & 53.1 & \\
    \multirow{2}{*}{DriveLaW~\citep{xia2026drivelaw}} & \multirow{2}{*}{CVPR'26} & S1 & 97.3 & 89.1 & 99.2 & 99.6 & 84.3 & \textbf{97.1} & 96.2 & \textbf{97.8} & 67.6 & \multirow{2}{*}{30.6} \\
    & & S2 & \textbf{82.5} & 67.6 & 83.5 & 98.1 & 84.8 & 78.5 & 45.8 & 96.4 & 57.3 & \\
    \midrule
    \rowcolor{defrow} & & S1 & \textbf{98.0} & \textbf{92.0} & \textbf{99.7} & 99.6 & 83.8 & 96.2 & \textbf{97.3} & \textbf{97.8} & 71.6 & \\
    \rowcolor{defrow} \multirow{-2}{*}{SimWAM (\textbf{ours})} & & S2 & 81.8 & \textbf{78.6} & \textbf{87.3} & 98.4 & 86.3 & \textbf{78.6} & \textbf{49.5} & 96.3 & 69.9 & \multirow{-2}{*}{\textbf{37.6}} \\
    \bottomrule
  \end{tabular}
  \vspace{-12pt}
\end{table}

\section{Additional Experimental Results}
\label{app:moreexp}

\subsection{NAVSIM-v2 Results}
\label{app:navsimv2}

We also evaluate SimWAM under NAVSIM-v2~\citep{Cao2025CORL} on both \texttt{navtest} and the safety-critical \texttt{navhard} benchmark, as shown in Tabs.~\ref{tab:navtest_v2} and~\ref{tab:navhard_v2}. Even before reinforcement learning, SimWAM attains the highest EPDMS of $90.2$ on \texttt{navtest} and $37.6$ on \texttt{navhard} among the compared methods, surpassing DriveLaW~\citep{xia2026drivelaw} by $1.6$ EPDMS points on \texttt{navtest} and by $7.0$ on \texttt{navhard}, with leading DAC, DDC, TTC, and LK, especially in the reactive second stage \begin{wrapfigure}{r}{0.45\textwidth}
  \centering
  \includegraphics[width=0.44\textwidth]{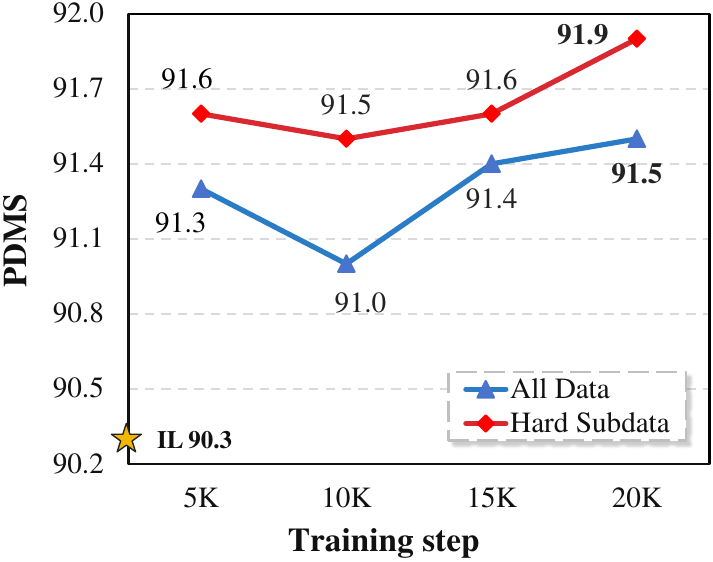}
  \caption{Ablation on the RL training data. The challenging subset with imitation PDMS below $90$ consistently outperforms the full \texttt{navtrain} set.}
  \label{fig:training_data}
  \vspace{-70pt}
\end{wrapfigure}where surrounding agents respond to the ego vehicle. Collectively, these results show that training-time world modeling delivers superior planning without any future-frame synthesis at inference. Moreover, the learned dynamics prior remains effective even when traffic reacts to the planner.

\begin{table}[t!]
  \centering
  \scriptsize
  \begin{minipage}[t]{0.49\linewidth}
    \centering
    \setlength{\tabcolsep}{2.6mm}
    \renewcommand{\arraystretch}{0.9}
    \captionof{table}{Ablation on future-video supervision.}
    \label{tab:ablation_frames}
    \scriptsize
    \vspace{-4pt}
    \begin{tabular}{lcccc Q}
      \toprule
      Setting & NC & DAC & EP & TTC & PDMS \\
      \midrule
      4\,f, 2\,s, 2\,Hz & 98.6 & 97.7 & 83.9 & 95.5 & 89.9 \\
      4\,f, 4\,s, 1\,Hz & \textbf{98.7} & 97.9 & \textbf{84.2} & 95.6 & 90.2 \\
      \rowcolor{defrow} 8\,f, 4\,s, 2\,Hz & 98.6 & \textbf{98.0} & 84.0 & \textbf{95.8} & \textbf{90.3} \\
      \bottomrule
    \end{tabular}
  \end{minipage}
  \hfill
  \begin{minipage}[t]{0.50\linewidth}
    \centering
    \setlength{\tabcolsep}{1.0mm}
    \renewcommand{\arraystretch}{0.9}
    \captionof{table}{The effect of input resolution.}
    \label{tab:ablation_res}
    \scriptsize
    \vspace{-4pt}
    \begin{tabular}{lcccc P c}
      \toprule
      Resolution & NC & DAC & EP & TTC & PDMS & Latency (ms)$\downarrow$ \\
      \midrule
      $192{\times}352$ & 98.2 & 97.1 & 83.0 & 94.9 & 88.9 & 509 \\
      \rowcolor{defrow} $384{\times}672$ & 98.6 & 98.0 & 84.0 & 95.8 & 90.3 & 518 \\
      $768{\times}1344$ & \textbf{98.7} & \textbf{98.1} & \textbf{84.3} & \textbf{96.1} & \textbf{90.6} & 573 \\
      \bottomrule
    \end{tabular}
  \end{minipage}
  \vspace{-10pt}
\end{table}

\begin{table}[t!]
  \centering
  \scriptsize
  \begin{minipage}[t]{0.49\linewidth}
    \centering
    \setlength{\tabcolsep}{1.7mm}
    \renewcommand{\arraystretch}{0.9}
    \captionof{table}{The effect of sampling steps.}
    \label{tab:ablation_steps}
    \scriptsize
    \vspace{-4pt}
    \begin{tabular}{lcccc P c}
      \toprule
      Steps & NC & DAC & EP & TTC & PDMS & Latency (ms)$\downarrow$ \\
      \midrule
      1 & 97.4 & 91.3 & 79.1 & 83.3 & 68.9 & 115 \\
      5 & \textbf{98.6} & 97.9 & \textbf{84.0} & 95.6 & 90.1 & 297 \\
      \rowcolor{defrow} 10 & \textbf{98.6} & \textbf{98.0} & \textbf{84.0} & \textbf{95.8} & \textbf{90.3} & 518 \\
      20 & \textbf{98.6} & \textbf{98.0} & 83.9 & \textbf{95.8} & 90.2 & 968 \\
      \bottomrule
    \end{tabular}
  \end{minipage}
  \hfill
  \begin{minipage}[t]{0.50\linewidth}
    \centering
    \setlength{\tabcolsep}{1.5mm}
    \renewcommand{\arraystretch}{0.9}
    \captionof{table}{Ablation of the RL scheme.}
    \label{tab:ablation_rl}
    \scriptsize
    \vspace{-4pt}
    \begin{tabular}{cc ccccc}
      \toprule
      Anchor Loss & $w$ & NC & DAC & EP & TTC & PDMS \\
      \midrule
      - & 0.2,1,1 & 94.7 & 94.4 & 85.1 & 87.5 & 84.3 \\
      \ding{51} & 1,1,1 & 98.5 & 99.0 & 85.5 & 96.2 & 91.6 \\
      \ding{51} & 0.5,1,1 & \textbf{98.6} & 99.0 & 85.7 & \textbf{96.4} & 91.8 \\
      \rowcolor{defrow} \ding{51} & 0.2,1,1 & 98.5 & \textbf{99.1} & \textbf{86.5} & 96.1 & \textbf{91.9} \\
      \bottomrule
    \end{tabular}
  \end{minipage}
  \vspace{-20pt}
\end{table}

\subsection{Additional Ablations}
\label{app:additional_results}

\noindent\textbf{Ablation on the prediction horizon and frame density.}
We examine the temporal configuration of future-video supervision in Tab.~\ref{tab:ablation_frames}. Shortening the prediction horizon from $4$ s to $2$ s noticeably reduces PDMS, whereas maintaining the $4$ s horizon with half as many frames recovers most of the performance. This comparison indicates that broad temporal coverage is more important than dense frame sampling for learning traffic dynamics. The full $4$ s target at $2$ Hz achieves the strongest performance.

\noindent\textbf{Ablation on the input resolution.}
We next study the trade-off between visual detail and inference efficiency in Tab.~\ref{tab:ablation_res}. Increasing the resolution from $192{\times}352$ to $384{\times}672$ substantially improves PDMS by $1.4$ points with only $9$ ms of additional latency. Further increasing the resolution to $768{\times}1344$ yields merely a $0.3$ point gain while adding considerably more computation. These results identify $384{\times}672$ as the most favorable balance between planning accuracy and inference efficiency.

\noindent\textbf{Ablation on the number of sampling steps.}
We further investigate the convergence of the action flow sampler in Tab.~\ref{tab:ablation_steps}. A single sampling step is insufficient to produce well-refined trajectories, whereas five steps already recover most of the performance. Increasing the budget to ten steps achieves the highest PDMS of $90.3$. Twenty steps provide no further improvement while nearly doubling the latency, indicating that performance saturates around ten steps in our setting.

\noindent\textbf{Ablation on the reinforcement-learning scheme.} Tab.~\ref{tab:ablation_rl} ablates the two design choices of our reinforcement-learning stage under a common training budget. Raising the longitudinal weight of the mean anchor to $0.5$ or $1.0$ slightly reduces PDMS, confirming that the \emph{down-weighted longitudinal coordinate} leaves the policy more freedom to optimize the driving reward. In contrast, removing the anchor loss collapses PDMS to $84.3$, even below the imitation checkpoint, showing that the anchor is \emph{essential} to stable policy updates rather than an optional regularizer.

\noindent\textbf{RL training data.} Fig.~\ref{fig:training_data} compares reinforcement learning on the full \texttt{navtrain} split and on the challenging subset whose imitation-policy PDMS is below $90$. The subset consistently outperforms the full set throughout training and reaches $91.9$ at $20$k steps. The difficult scenarios expose clearer differences among the sampled trajectories and thus provide more informative reward signals. In contrast, many scenes in the full set are already well handled by imitation learning and contribute limited learning signals, which motivates our subset restriction.

\section{More Qualitative Results}
\label{app:qualitative}

\subsection{Future-Video and Trajectory Visualization}

\begin{figure}[t!]
  \centering
  \includegraphics[width=0.98\textwidth]{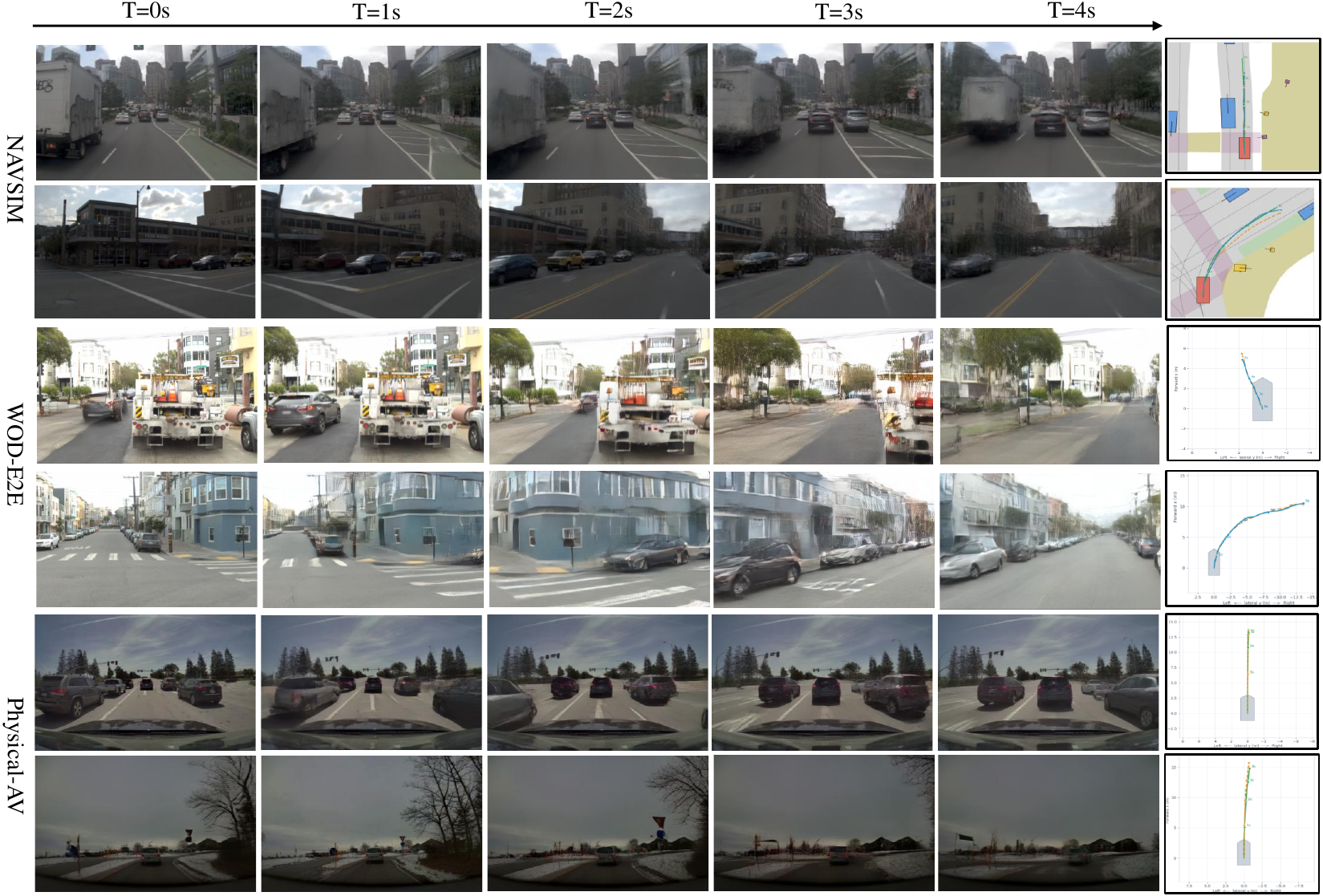}
  \caption{Additional qualitative results of SimWAM across NAVSIM, WOD-E2E, and PhysicalAI-AV, covering diverse streets, intersection turns, large leading trucks, and snow-covered roads.}
  \label{fig:qualitative_more}
  \vspace{-10pt}
\end{figure}

As shown in Fig.~\ref{fig:qualitative_more}, we present qualitative results on more diverse scenarios, spanning dense urban streets, left and right turns at intersections, close following of large trucks, and snow-covered roads across the three benchmarks. In each case, SimWAM anticipates the evolution of the surrounding traffic over the four-second horizon, and the predicted video stream remains consistent with the planned trajectory shown in the right column. Notably, the snow-covered example presents challenging visual conditions, while the predicted trajectory remains visually smooth. These examples corroborate the coherent and reliable behavior of SimWAM under varied driving conditions.

\clearpage

\subsection{Trajectory Visualization before and after Reinforcement Learning}

\begin{figure}[t!]
  \centering
  \includegraphics[width=0.98\textwidth]{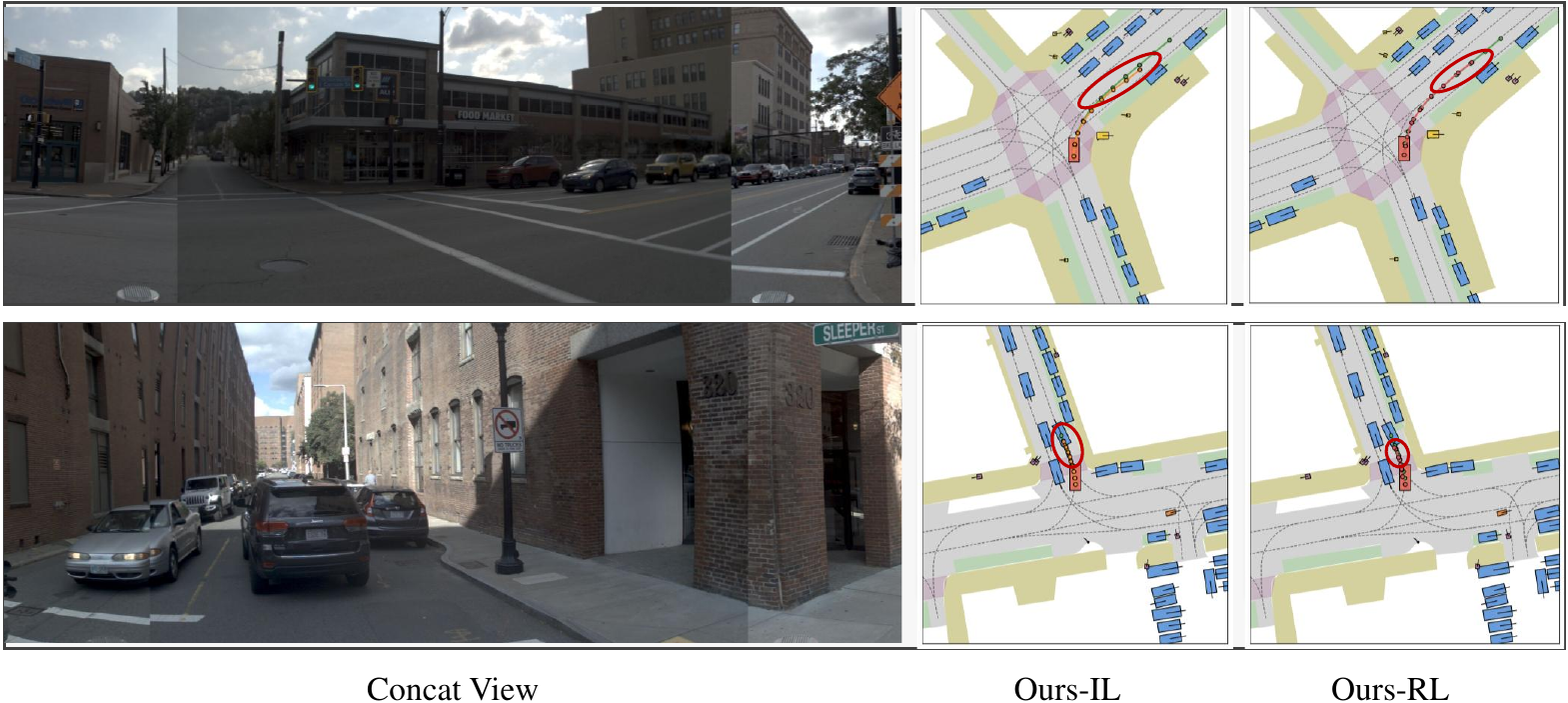}
  \caption{Qualitative comparison of \emph{Ours-IL} and \emph{Ours-RL} on two \texttt{navtest} scenarios. Red ellipses highlight regions where \emph{Ours-RL} progresses farther while remaining within the drivable area.}
  \label{fig:qualitative_rl}
  \vspace{-10pt}
\end{figure}

Fig.~\ref{fig:qualitative_rl} compares the imitation-trained and RL-fine-tuned models on two scenes. The imitation-trained model produces conservative trajectories and advances only a short distance at the intersection and along the narrow street. After reinforcement learning, the model follows the intended route more decisively and completes a larger portion of each maneuver, while the trajectories remain within the drivable area and keep a safe clearance from surrounding vehicles. This comparison shows that the driving reward mitigates the conservative predictions of the imitation-trained policy.

\end{document}